\def\year{2020}\relax
\documentclass[letterpaper]{article} 
\usepackage{aaai20}  
\usepackage{times}  
\usepackage{helvet} 
\usepackage{courier}  
\usepackage[hyphens]{url}  
\usepackage{graphicx} 
\usepackage{graphicx}  
\usepackage{amsfonts}       
\usepackage{nicefrac}       
\usepackage{microtype}      

\usepackage{epsfig}
\usepackage{amsmath}
\usepackage{amssymb}
\usepackage{epstopdf}
\usepackage{color}
\usepackage{array}
\usepackage{comment}
\usepackage{algorithm}
\usepackage{algorithmicx}
\usepackage{cases}
\usepackage{algpseudocode}
\usepackage{multirow}
\usepackage{textcomp}
\usepackage{subfigure}
\usepackage{booktabs}

\usepackage{subfig}
\usepackage{booktabs}
\usepackage[export]{adjustbox}

\title{Separate In Latent Space: Unsupervised Single Image Layer Separation}
\author{Yunfei Liu, ~~~Feng Lu\thanks{Corresponding Author} \\
	State Key Laboratory of Virtual Reality Technology and Systems,\\
	School of Computer Science and Engineering, Beihang University, Beijing, China  \\
	\texttt{\{lyunfei, lufeng\}@buaa.edu.cn}}
\begin{document}
	
	\maketitle
	
	\newcommand{\todo}[1]{{\color{red}{ \textbf{TODO: #1}}}}
	\newcommand{\lf}[1]{{\color{magenta}{#1}}}
	\newcommand{\yunfei}[1]{{\color{cyan}{\textbf{#1}}}}
	\newcommand{\eg}{\textit{e.g.}}
	\newcommand{\ie}{\textit{i.e.}}
	\newcommand{\etal}{\textit{et al}}
	\newcommand{\etc}{\textit{etc}}
	\newcommand{\tabincell}[2]{\begin{tabular}{@{}#1@{}}#2\end{tabular}}
	
	\begin{abstract}
		Many real world vision tasks, such as reflection removal from a transparent surface and intrinsic image decomposition, can be modeled as single image layer separation. However, this problem is highly ill-posed, requiring accurately aligned and hard to collect triplet data to train the CNN models. To address this problem, this paper proposes an unsupervised method that requires no ground truth data triplet in training. At the core of the method are two assumptions about data distributions in the latent spaces of different layers, based on which a novel unsupervised layer separation pipeline can be derived. Then the method can be constructed based on the GANs framework with self-supervision and cycle consistency constraints, \etc. Experimental results demonstrate its successfulness in outperforming existing unsupervised methods in both synthetic and real world tasks. The method also shows its ability to solve a more challenging multi-layer separation task.
	\end{abstract}
	
	
	\section{Introduction} \label{sec:intro}
	
	Real imaging process is influenced by different factors, \eg, shape, illumination, reflection and refraction, making the captured images highly complex. Such images, when used as input data to an AI system, may heavily affect the system performance on various vision-based tasks, such as object detection and 3D reconstruction. As a result, separating different visual attributes from a single image has been considered an important research goal for decades.
	
\begin{figure} [htbp]
		\centering
		\includegraphics[width=\linewidth]{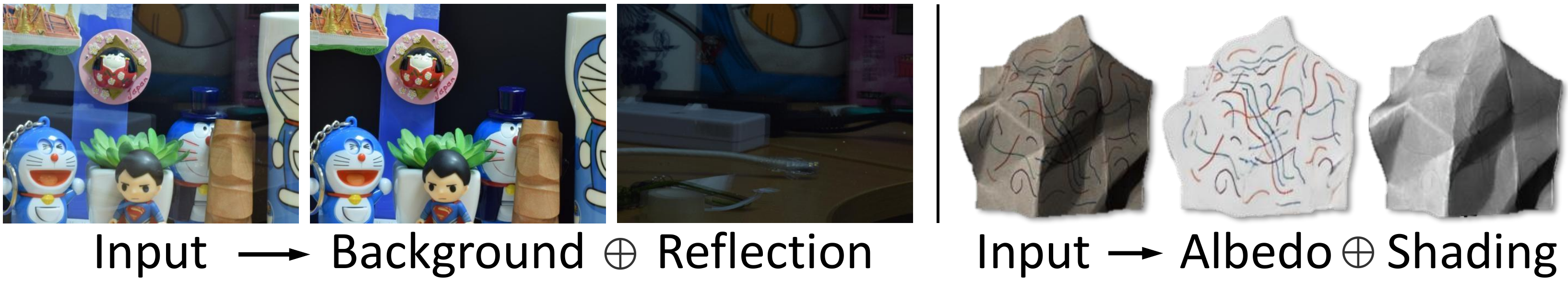}
		\caption{Examples of single image layer separation: reflection separation and intrinsic image decomposition. }
		\label{fig:rils_samples}
	\end{figure}

	This paper considers visual attributes separation via Single Image Layer Separation (SILS).
	It aims at decomposing an image into several independent layers with specific physical meanings. In general, this problem can be formulated as:
	\begin{equation} \label{equ:intro}
	I = L_1 \oplus  L_2 \oplus  \cdots,
	\end{equation}
	where $L_i$ is the $i$-th layer and $I$ is the input image that can be reconstructed by a pixel-wise blend of all the layers. Plenty of real world problems can be represented in this form.
	As shown in Fig.~\ref{fig:rils_samples}, for {reflection separation}, the input image $I$ can be decomposed into a reflection layer $L_R$ and a background layer $L_B$ in the form of $I=L_R+L_B$. While for {intrinsic decomposition}, the input image $I$ can be represented by the pixel-wise product of an albedo image $L_A$ and a shading image $L_S$, or alternatively formulated as $\log(I)= \log(L_A)+\log(L_S)$.
	
	Despite its wide applicability, single image layer separation has been found fundamentally ill-posed, due to the existence of countless feasible solutions. To constrain the solution space, conventional methods use different priors such as low rank~\cite{PPP:han2017reflection_low_rank}, sparsity~\cite{PPP:li2013exploiting_sparsity} and relative smoothness~\cite{PPP:Li2014Single} in optimization. Such hand-crafted constraints may work well in specific cases but fail in more complex cases.
	Recently,
	the deep convolutional neural network (CNN) has been widely used to handle similar problems~\cite{PPP:Fan2018revisitingDeepIntrinsic,PPP:Yang18BDN,PPP:Zhang2018PerceptualLosses}. However, CNN-based methods face major difficulties in high-quality data acquisition:
	1) it is difficult or even impossible to obtain the ground truth of each layer for real images, and 2) different datasets vary greatly and cannot be used jointly. Since the performance of CNNs relies heavily on the training data, existing methods have to either use synthetic data with limited reality or train their networks with imperfect data.
	
	In order to overcome such difficulties, this paper suggests \emph{not requiring the ground truth data triplet}, \ie, one input image and its exact two layers as training data, but simply collecting images from the three domains independently. In this manner, sufficient training data can be easily obtained without the acquisition problem. The absence of ground truth triplet, however, makes it an \emph{unsupervised learning problem}. Therefore, we propose an unsupervised single image layer separation (unsupervised SILS) method in this paper.
	
	The proposed unsupervised SILS method does not rely on training data triplets from the three domains. Instead,
	we introduce two key assumptions, namely the shared information consistency and the layer independence, to help link the three different domains and construct the latent spaces. Based on the assumptions, we design the method by using techniques such as self-supervision, cycle consistency and GANs. The full method is able to perform unsupervised SILS without requiring ground truth triplets in training.
	
	We test our method on two classical tasks, \ie, single image reflection separation and intrinsic image decomposition, and it outperforms state-of-the-art unsupervised methods. As our method does not need ground truth triplets, it can be quickly applied to other layer separation tasks with little data collection burden. Finally, our method can be extended to solve the more challenging multi-layer separation task.
	
	
	\subsection{Related Works}
	The problem of image layer separation can be considered as a special case of image domain translation. Previous domain translation can be formulated as a mapping function between a source and a target domain~\cite{UII:Radford2016DCGAN,UII:Yi2017Dualgan,UII:Zhu2017cycleGAN,UII:Liu2017UnsupervisedImageTranslation}, while single image separation is a mapping from the source domain to two other domains physically related to the source domain. There are various image separation problems in the field of computer vision where different physical models are applied as priors for separation tasks. For instance, relative smoothness~\cite{PPP:Li2014Single}, ghost cues~\cite{PPP:Shih2015Reflection} and layer independence priors~\cite{PPP:li2013exploiting_sparsity}
	are introduced for separating and removing the reflection on the glass surface from the background.
	However, the above mentioned methods can only handle simple cases with image gradient or color changes, and can not be adopted to more complex situations.
	
	In recent years, supervised deep learning methods are applied to many image separation tasks. For example, fully convolutional networks 
	with various guidance, \eg, image gradient  information~\cite{PPP:Fan2017CEILNet}, face structure priors~\cite{PPP:Wan2019FaceReflection} and perceptual losses~\cite{PPP:Zhang2018PerceptualLosses}, are used for single image reflection removal. Besides, Ronneberger \etal. used U-net-like CNNs for intrinsic image decomposition~\cite{IU:Ronneberger2015Unet}.
	
	Since real shot images with layer separation results are hard to collect, the development of unsupervised algorithms is in great need.
	Michael \etal~\cite{UII:Michael2017SelfSupervisedIntrinsit} proposed a self-supervised intrinsic image decomposition method. It can be trained with only a few images with ground-truth, and then transferred to other unpaired images. However, the albedo layer should be the same among different training images.
	Li \etal~\cite{PPP:Li2018BigTime} and Lettry~\etal~\cite{PPP:LVvG18} proposed unsupervised intrinsic image decomposition methods, but these methods need multiple inputs with the same albedo layer for training.
	More recently, Hoshen~\etal~\cite{UII:hoshen2019towards} proposed an unsupervised single-channel blind source separation method, while the method requires a pair of input images which limits its applicability.
	Overall, unsupervised single image layer separation remains an open problem to solve.

	
	\section{Problem Formulation and Analysis}\label{sec:problemFormulation}
	
	\begin{figure*} [ht]
		\begin{center}
			\includegraphics[width=0.7\textwidth]{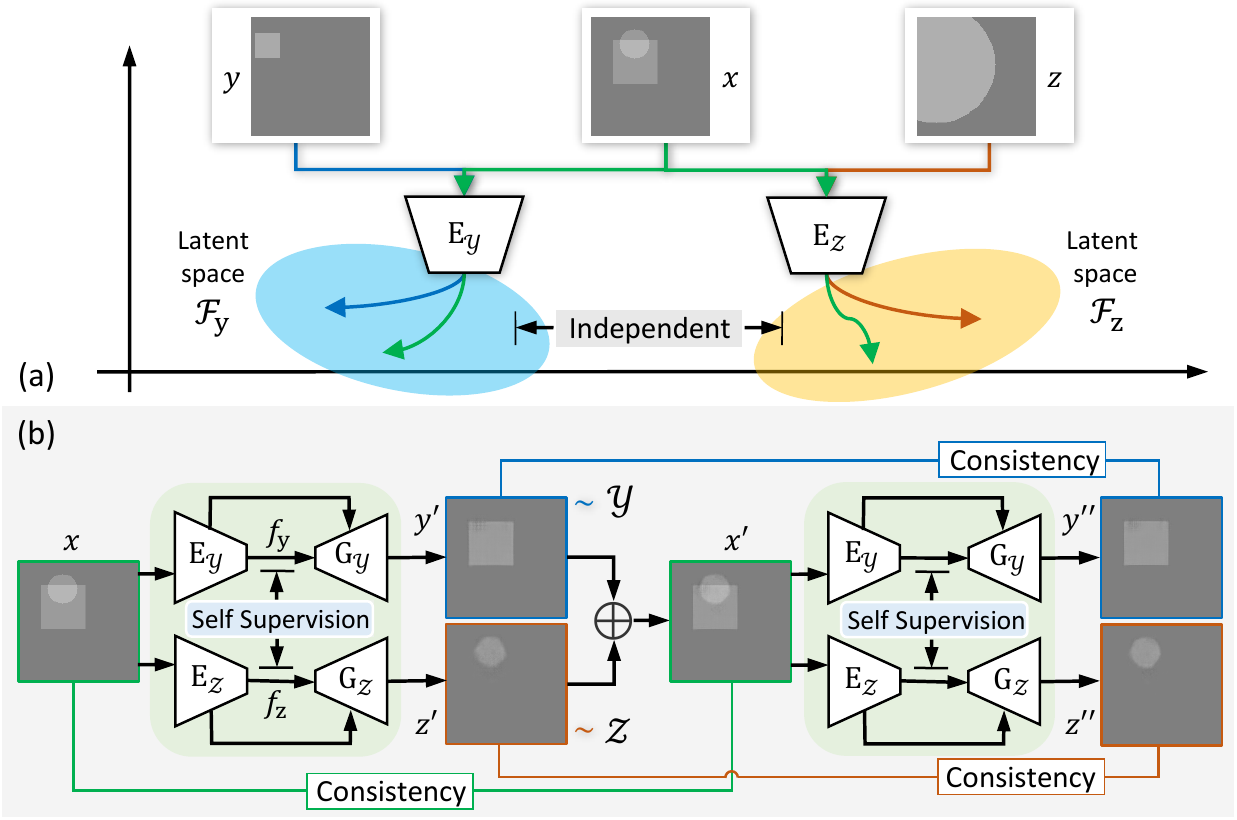}
		\end{center}
		\caption{ (a) Illustrations of the shared information consistency and layer independence assumptions. $E_\mathcal{Y}$, $E_\mathcal{Z}$ are two encoding functions that map images into different and independent latent spaces ($\mathcal{F}_y$ for square and $\mathcal{F}_x$ for circle).
			Intuitively speaking, the square features in both $x$ and $y$ go to the latent space $\mathcal{F}_y$, and the circle features in both $x$ and $z$ go to the latent space $\mathcal{F}_z$. In this manner, the method learns how to separate information from $x$.
			(b) The overall framework of our unsupervised SILS method.
			The encoder/decoder networks $E_\mathcal{Y}$, $E_\mathcal{Z}$, $G_\mathcal{Y}$ and $G_\mathcal{Z}$ are implemented by using CNNs.
			Following our two assumptions, we utilize the cycle consistency constraint, and also the self-supervision learning.
			The $\sim$ indicates where we add adversarial discriminators for evaluating whether the generated images are realistic.
		}
		\label{fig:main_figure}
	\end{figure*}
	
	Single Image Layer Separation can be considered as a one-to-two image domain translation problem. In particular,
	let $\mathcal{X}, \mathcal{Y}$ and $\mathcal{Z}$ denote three image domains, then we try to find out a proper separation $x = y \oplus z$.
	In the supervised case, we have training samples $(x, y, z)$ drawn from a joint distribution $P_{\mathcal{X}, \mathcal{Y}, \mathcal{Z}}(x, y, z)$.
	While in the unsupervised case, we only have samples drawn from the marginal distributions $P_{\mathcal{X}}(x)$,  $P_{\mathcal{Y}}(y)$ and $P_{\mathcal{Z}}(z)$.
	Since we can infer nothing about the joint distribution only from the marginal samples without additional assumptions, the unsupervised problem is fundamentally ill-posed, as mentioned in Sec.~\ref{sec:intro}. To address this problem, we introduce the following assumptions.

	\textbf{Assumption A: Shared information consistency.}
	Given that $x =y \oplus z$, we assume that there exists a shared latent space $\mathcal{F}_y$ for $x$ and $y$, \ie, $\forall x,y$, $E_{\mathcal{Y}}(x)\in \mathcal{F}_y$ and $E_{\mathcal{Y}}(y)\in \mathcal{F}_y$, where $E_{\mathcal{Y}}$ is a function that maps data from the color image space to the latent space $\mathcal{F}_y$. Similarly, we have $E_{\mathcal{Z}}(x)$ and $E_{\mathcal{Z}}(z)$ both belong to the same latent space $\mathcal{F}_z$.
	
	Following this assumption (also see Fig.~\ref{fig:main_figure}~(a)), the original image layer separation pipeline can be reformed as:
	
	\begin{equation} \label{equ:assump1}
	x \to \{y, z\}
	\Rightarrow
	x \begin{array}{l}{\nearrow} \\ {\searrow}\end{array}
	\begin{array}{l}{f_y \stackrel{G_{\mathcal{Y}}}{\longrightarrow} y} \\  \\
	{f_z \stackrel{G_{\mathcal{Z}}}{\longrightarrow} z}\end{array},
	\end{equation}
	where $f_y = E_{\mathcal{Y}}(x)\in \mathcal{F}_y$ and $f_z = E_{\mathcal{Z}}(x)\in \mathcal{F}_z$.
	Such an $f_y$ can be understood as the extracted information, which only belongs to the domain $\mathcal{Y}$, from the blended input $x$.
	Then, $G_{\mathcal{Y}}$ is another mapping function to project $f_y$ back to the color image space and generate the layer image $y$.\
	
	In conclusion, this assumption allows producing $y/z$ from a single $x$ if we find out correct $E_{\mathcal{Y}}/E_{\mathcal{Z}}$ and $G_{\mathcal{Y}}/G_{\mathcal{Z}}$.
	
	\textbf{Assumption B: Layer independence.}
	We assume that the separated $y$ and $z$ should be independent to each other in the latent space, \ie, samples belonging to the same domain have more similar features in the latent space.
	
	Computationally, for any two samples $f_y^a$, $f_y^b$ from $\mathcal{F}_y$, and any other two samples $f_z^a$, $f_z^b$ from $\mathcal{F}_z$, we assume:
	
	\begin{equation} \label{equ:assump2}
	\begin{aligned}
	d_\psi \gg d_\phi, \forall d_\psi \in \{|f_y^a - f_z^a|, |f_y^a - f_z^b|, |f_y^b - f_z^a|,\\ |f_y^b - f_z^b|\} \; and\;
	\forall d_\phi \in \{ |f_y^a - f_y^b|, |f_z^a - f_z^b| \},
	\end{aligned}
	\end{equation}
	where $d_\psi$ indicates an inter latent space distance and $d_\phi$ indicates an inner latent space distance.
	This assumption can be satisfied by minimizing $d_\phi$ while maximizing $d_\psi$.

	
	\section{Learning for Unsupervised SILS}
	
	Based on the two assumptions in Sec.~\ref{sec:problemFormulation}, we propose an unsupervised single image layer separation method. In particular, our method implements self-supervision, cycle consistency, GANs, \etc. to achieve efficient unsupervised learning. The overall framework of our method can be seen in Fig.~\ref{fig:main_figure}~(b).
	
	\textbf{Self Supervision (SS).}
	According to assumption A, $x$ contains all the information to produce $f_y$ and $f_z$, respectively. Meanwhile, assumption B requires that $f_y$ and $f_z$ should be as different as possible. Therefore, we need to carefully design the encoders $E_\mathcal{Y}(x)$ and $E_\mathcal{Z}(x)$ so that their outputs have maximized distances.
	
	To this end, we propose a novel self-supervision framework to optimize both encoders and constrain the feature distributions in both latent spaces.
	In particular, we maximize the distance between output features from $E_\mathcal{Y}$ and $E_\mathcal{Z}$;
	for $E_\mathcal{Y}$, we minimize the $L1$ distance between $E_\mathcal{Y}(x_i)$ and $E_\mathcal{Y}(y_i)$, where $x_i$ and $y_i$ are unpaired samples randomly selected from $\mathcal{X}$ and $\mathcal{Y}$;
	for $E_\mathcal{Z}$, we minimize the $L1$ distance between $E_\mathcal{Y}(x_j)$ and $E_\mathcal{Z}(z_j)$, where $x_j$ and $z_j$ are unpaired samples randomly selected from $\mathcal{X}$ and $\mathcal{Z}$.
	
	\textbf{Cycle Consistency (CC).}
	Assumption A describes how to produce $y$ and $z$ from the input $x$, while there should be one more step to close the loop: that $y$ and $x$ should re-produce $x$. This is physically obvious and leads to the cycle consistency.
	
	As shown in Fig.~\ref{fig:main_figure}~(b), let $y'$ and $z'$ be the output of $G_{\mathcal{Y}}$ and $G_{\mathcal{Z}}$, then we can simply fuse them again to produce $x' = y' \oplus z'$. Here we introduce the first cycle consistency constraint that enforces $x=x'$. Moreover, the new $x'$ can be further separated into new $y''$ and $z''$, resulting the second cycle consistency constraint of $y'=y''$ and $z'=z''$. In this manner, the physical correctness of the image layer separation \& fusion processes is guaranteed.

	\textbf{GANs.}
	Assumption A and B mainly focus on the separability of different layers, while there is no guarantee that the separated images look realistic.
	Therefore, we adopt the widely successful Generative Adversarial Networks (GANs) in our framework. The aim is to ensure the output $y$ and $z$ show realistic visual appearances as real images in their respective domains $\mathcal{Y}$ and $\mathcal{Z}$.
	
	The GANs structure comprises generators and discriminators. For the former, we have $E_\mathcal{Y} \circ G_\mathcal{Y}$ and $E_\mathcal{Z} \circ G_\mathcal{Z}$ work as the generator to produce layer images $y$ and $z$, as shown in Fig.~\ref{fig:main_figure}~(b).
	As for the latter, we add two discriminators $D_\mathcal{Y}$ and $D_\mathcal{Z}$. They are used to distinguish the outputs of the generators from the real images in domains $\mathcal{Y}$ and $\mathcal{Z}$.

\begin{figure*} [ht]
		\centering
		
		\subfigure[]{
			\adjustbox{valign=b}{
				\includegraphics[width=0.276\linewidth, valign=b]{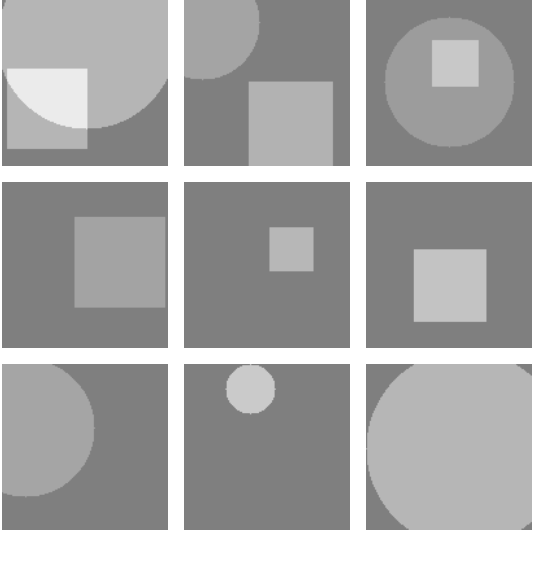}}}
		\subfigure[]{
			\adjustbox{valign=b}{
				\includegraphics[width=0.3\linewidth, valign=b]{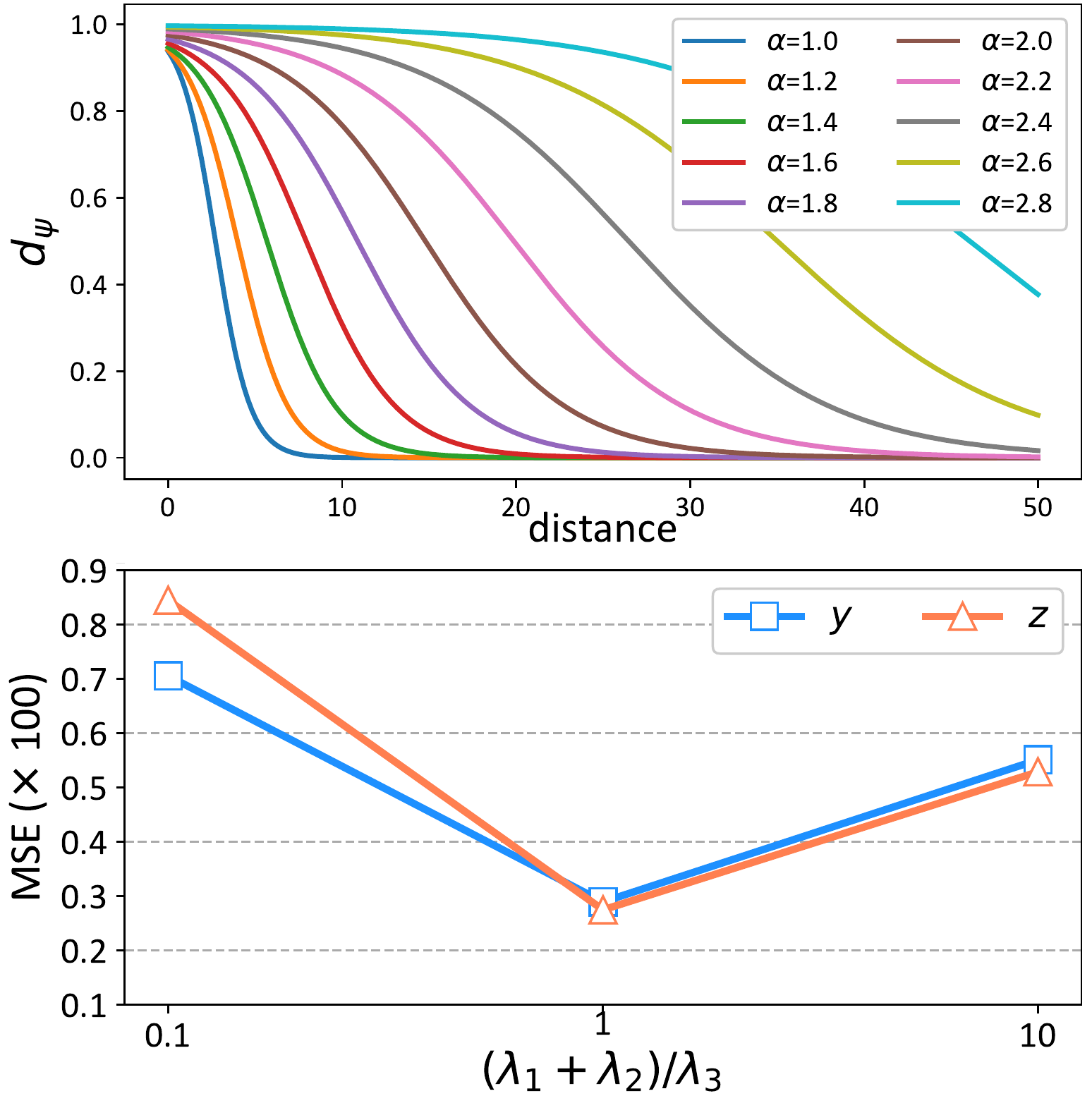}}}\quad
		\subfigure[]{
			\adjustbox{valign=b}{
				\begin{tabular}{ccc}
					\toprule[1.3pt]
					\specialrule{0em}{1pt}{7pt}
					Method     &  $y'$    & $z'$    \\
					\specialrule{0em}{1pt}{6pt}
					\hline
					\specialrule{0em}{1pt}{7pt}
					CycleGAN       &    0.671      &  0.610   \\
					\specialrule{0em}{1pt}{6pt}
					UNIT       &     0.463     &  0.372   \\
					\specialrule{0em}{1pt}{6pt}
					\hline
					\specialrule{0em}{1pt}{7pt}
					Ours w/o CC     &  0.717   & 0.715   \\
					\specialrule{0em}{1pt}{7pt}
					Ours w/o SS     &  0.604   & 0.548   \\
					\specialrule{0em}{1pt}{7pt}
					Ours full   &  \bf{0.289}   & \bf{0.274}   \\
					\specialrule{0em}{1pt}{7pt}
					\bottomrule[1.3pt]
					\specialrule{0em}{1pt}{12pt}
		\end{tabular}}}
		
		\caption{(a) Demonstration of the synthetic non-triplet data. Top row: input image $x$ (a blend of circle and square), middle row: $y$ (only square), and bottom row: $z$ (only circle).
			Our purpose is to decompose $x$ into $y\oplus z$ by learning from these data.
			(b) Visualization following the analysis in Sec.~\ref{sec:toy_exp_and_perform_analysis}. Top: distance loss curve
			v.s. different $\alpha$. Bottom: separation performance v.s. different hyper parameters in SS.
			(c) Illustration of how self-supervision (SS) and cycle consistency (CC) affect the performance, and a comparison with existing unsupervised methods. Note that the numbers indicate image MSE $\times$ 100.}
		\label{fig:performance_analysis}
	\end{figure*}

	\textbf{Loss Function Design.}
	According to the techniques introduced above, we propose the detailed loss functions as following. The overall loss function is designed as:
	\begin{equation} \label{equ:total_object_func}
	\begin{split}
	\min_{E, G}\max_{D}  \mathcal{L}_{SS}(E_\mathcal{Y}, E_\mathcal{Z}) + \mathcal{L}_{GAN_\mathcal{Y}}(E_\mathcal{Y}, G_\mathcal{Y}, D_\mathcal{Y}) \\
	+ \mathcal{L}_{GAN_\mathcal{Z}}(E_\mathcal{Z}, G_\mathcal{Z}, D_\mathcal{Z})
	+ \mathcal{L}_{CC_\mathcal{X}}(E_\mathcal{Y}, G_\mathcal{Y}, E_\mathcal{Z}, G_\mathcal{Z}) \\
	+ \mathcal{L}_{CC_\mathcal{Y}}(E_\mathcal{Y}, G_\mathcal{Y})
	+ \mathcal{L}_{CC_\mathcal{Z}}(E_\mathcal{Z}, G_\mathcal{Z}).
	\end{split}
	\end{equation}
	The first part is related to the self-supervised training.
	Based on Equ.~\eqref{equ:assump2}, the objective function can be designed as:
	\begin{equation} \label{equ:ss_object_func}
	\begin{split}
	\mathcal{L}_{SS}(E_\mathcal{Y}, E_\mathcal{Z}) &= \lambda_1 d_\phi (E_\mathcal{Y}(y), E_\mathcal{Y}(x)) \\
	&+ \lambda_2 d_\phi (E_\mathcal{Z}(z), E_\mathcal{Z}(x)) \\
	&+ \lambda_3 (1-d_\psi (E_\mathcal{Y}(x), E_\mathcal{Z}(x))),
	\end{split}
	\end{equation}
	where the hyper-parameters $\lambda_1$, $\lambda_2$ and $\lambda_3$ control the weights of different objective terms.
	We adopt $d_\phi(a, b) = |a - b|$ in practice, \ie, we compute the $L1$ distance of two inputs $a, b$.
	Besides, $d_\psi$ measures the distance between the latent codes which are the outputs of $E_\mathcal{Y}$ and $E_\mathcal{Z}$, and we use a modified sigmoid function to compute the distance:
	\begin{equation} \label{equ:loss_distance_of_latent}
	d_\psi (a, b) = \frac{1}{1 + e^{g(a, b)}}, \text{ where } g(a, b) = -\frac{|a - b| - \alpha e^\alpha}{\alpha^2}.
	\end{equation}
	Here, $\alpha$ controls the shape of the distance curve. Its effect is shown by the experimental results in the top of Fig.~\ref{fig:performance_analysis} (b).
	
	The GANs-related objective functions are as below:
	\begin{equation} \label{equ:gan_loss_func}
	\begin{split}
	\mathcal{L}_{GAN_\mathcal{Y}}(E_\mathcal{Y}, G_\mathcal{Y}, D_\mathcal{Y}) = \lambda_0 \mathbb{E}_{y\sim P_\mathcal{Y}}[\log D_\mathcal{Y}(y)] \\
	+ \lambda_0\mathbb{E}_{f_y\sim E_\mathcal{Y}(f_y|x)}[\log (1 -  D_\mathcal{Y}(G_\mathcal{Y}(f_y)))], \\
	\mathcal{L}_{GAN_\mathcal{Z}}(E_\mathcal{Z}, G_\mathcal{Z}, D_\mathcal{Z}) = \lambda_0 \mathbb{E}_{z\sim P_\mathcal{Z}}[\log D_\mathcal{Z}(z)] \\
	+ \lambda_0\mathbb{E}_{f_z\sim E_\mathcal{Z}(f_z|x)}[\log (1 - D_\mathcal{Z}(G_\mathcal{Z}(f_z)))].
	\end{split}
	\end{equation}
	
	The above functions are in the form of conditional GANs objective functions.
	The only hyper-parameter $\lambda_0$ controls the impact of the entire GAN objective functions.
	
	Finally, for the cycle-consistency constraint, the loss functions are based on the $L1$ differences:
	\begin{equation} \label{equ:consistency_loss_func}
	\begin{split}
	\mathcal{L}_{CC_\mathcal{X}} &= \lambda_4|y' + z' - x|, \\
	\mathcal{L}_{CC_\mathcal{Y}} &= \lambda_5|G_\mathcal{Y}(E_\mathcal{Y}(y'+z')) - y'|, \\	\mathcal{L}_{CC_\mathcal{Z}} &= \lambda_6|G_\mathcal{Z}(E_\mathcal{Z}(y'+z')) - z'|,
	\end{split}
	\end{equation}
	where $y' = G_\mathcal{Y}(E_\mathcal{Y}(x))$ and $z' = G_\mathcal{Z}(E_\mathcal{Z}(x))$.
	The hyper-parameters $\lambda_4$, $\lambda_5$ and $\lambda_6$ control the weights of these terms.
	
	\section{Implementation via CNNs}
	
	Our unsupervised SILS network consists of a two-stream generator and two discriminators, as shown in Fig.~\ref{fig:main_figure}~(b).
	The generator is composed of two encoder/decoder pairs, \ie, $E_\mathcal{Y} \circ G_\mathcal{Y}$ and $E_\mathcal{Z} \circ G_\mathcal{Z}$, to generate $y$ and $z$ from a single $x$.
	The discriminators determine whether the generated $y$ and $z$ look real enough.
	Below are the implementation details.
	
	Due to the different tasks in our experiments, we design two types of generators in practice.

	\textbf{Generator for synthetic data.}
	The synthetic data is rendered by using simple square and circle shapes, as shown in Fig.~\ref{fig:performance_analysis}~(a).
	Therefore, the generator is designed to be relatively simple.
	The structures of $E_\mathcal{Y} \circ G_\mathcal{Y}$ and $E_\mathcal{Z} \circ G_\mathcal{Z}$ are the same, but they do not share parameters.
	In detail, each of the encoders $E_\mathcal{Y}$ and $E_\mathcal{Z}$ has five convolutional layers, which have $\{16, 32, 64, 128, 256\}$ filters of the size $4\times4$ and a stride of $2$.
	Batch normalization~\cite{IU:Sergey2015BatchNorm} and leaky ReLU activation are applied after every convolutional layer.
	Each of the decoders $G_\mathcal{Y}$ and $G_\mathcal{Z}$ has the same layers as the encoder but in a reverse order plus a final layer with $3$ channels.
	The corresponding layers of the encoder and decoder have mirror-link connections as described in~\cite{IU:Ronneberger2015Unet}, yielding sharper results.
	
	\textbf{Generator for real image separation tasks.}
	In order to handle real world tasks, we propose a more complex generator as an alternative.
	Overall, $E_{\mathcal{Y}} \circ G_{\mathcal{Y}}$ and $E_{\mathcal{Z}} \circ G_{\mathcal{Z}}$ still share the same structure, but not the parameters.
	In detail, the encoders $E_{\mathcal{Y}}$ and $E_{\mathcal{Y}}$ are implemented by using VGG-19~\cite{ML:Simonyan2015VGG}. The decoders $G_\mathcal{Y}$ and $G_\mathcal{Z}$ have $4$ convolution blocks in each, and each block comprises a convolution layer, a leaky ReLU activation layer and an up-sample layer. The last block is followed by a fully convolutional network with $64$ filters of size 3$\times$3, a stride of $1$ and dilation rations of $\{2, 4, 8, 16, 32, 1\}$. Instance normalization~\cite{IU:Dmitry2015InstanceNorm} and leaky ReLU activation are applied after each convolution layer. The output layer is a convolution layer with three $1\times1$ filters.
	
	For the skip connection between the encoder and decoder, we select `conv1\_2', `conv2\_2' and `conv3\_2' layers in the encoder and make skip connections to the first block of the decoder. This strategy has been shown efficient for image synthesis and enhancement~\cite{PPP:Zhang2018PerceptualLosses}.

	\textbf{Discriminator.}
	We implement $D_\mathcal{Y}$ and $D_\mathcal{Z}$ as the discriminators.
	They share the same multi-branches structures as in~\cite{UII:Liu2017UnsupervisedImageTranslation}, but do not share parameters.
	For the $i$-th branch in $D_\mathcal{Y}$ or $D_\mathcal{Z}$, the input image is down-sampled by $i-1$ times via average pooling and then fed to the network.
	Each branch has four convolution layers with $\{32, 64, 128, 32\}$ filters of size $4\times4$ and a stride of $2$.
	In the end, the output features from different branches are fused together followed by a sigmoid activation.
	For the synthetic data case, the number of branches is simply $1$, while for the real image separation tasks, the number of branches is set to $3$.

	\textbf{Training.}
	We use ADAM~\cite{ML:Kingma2015Adam} optimizer with a learning rate of $0.0001$ and
	momentums of $0.0$ and $0.9$.
	The default values of hyper-parameters in Equ.~\eqref{equ:ss_object_func}-\eqref{equ:consistency_loss_func} are set to $\lambda_0 = 5.0$, $\lambda_1 = 0.5$, $\lambda_2 = 0.5$, $\lambda_3 = 1.0$, $\lambda_4 = 1.0$, $ \lambda_5 = 1.0$, $\lambda_6 = 1.0$ and $ \alpha = 1.4$.
	
	Each mini-batch in training contains one input image from the domain $\mathcal{X}$, one layer image from the domain $\mathcal{Y}$ and another one from domain $\mathcal{Z}$.
	Note that these three images do not form an image separation triplet.
	For the real world tasks, data argumentation operations are applied, including random scalings by factors within $0.8\thicksim1.2$, random croppings and random horizontal flippings.
	
	Inheriting from GANs, the training solves a min-max problem
	to find a saddle point in the solution space.
	To make it stable, we use a gradient update strategy similar to~\cite{UII:Liu2017UnsupervisedImageTranslation} with gradient penalization to solve Equ.~\eqref{equ:total_object_func}.
	Specifically, a gradient descent step first updates $E_\mathcal{Y}$, $E_\mathcal{Z}$, $G_\mathcal{Y}$ and $G_\mathcal{Z}$ with fixed $D_\mathcal{Y}$ and $D_\mathcal{Z}$, then another gradient descent step updates $D_\mathcal{Y}$ and $D_\mathcal{Z}$ with fixed $E_\mathcal{Y}$, $E_\mathcal{Z}$, $G_\mathcal{Y}$ and $G_\mathcal{Z}$.
	
	
	\section{Experiments}
	
	Experiments are conducted on the synthetic data and also real world single image separation tasks.
	The former aims at providing with intuitive and quantitative results, and the latter demonstrates the real world applicability of our method.
	Moreover, our method is extended to solve a more challenging task that involves multiple layers.
	
	\subsection{Datasets}
	We prepare three types of dataset:
	1) synthetic dataset, 2) real world intrinsic image decomposition dataset and 3) real world reflection removal dataset. Note that in any case, we do not have the data triplet from the three domains, \ie, input image with its two layers, in the training set.
	
	Examples of the synthetic data is shown in Fig.~\ref{fig:performance_analysis}~(a). The image size is set to $128\times128$. The squares and circles in the images are rendered with arbitrary sizes, positions and brightness levels. The training set includes $4000$ images and the test set has $1000$ images.
	
	For intrinsic image decomposition, we use $220$ images in the MIT intrinsic dataset~\cite{PPP:Grosse2011MITintrinsic}.
	To avoid the ground truth triplet in training, we divide the dataset into two subsets without overlapped scenes. Then we select the input images only from the first subset, and the layer images only from the second. The test samples are from the second subset.
	For reflection removal, we use the reflection removal benchmark dataset~\cite{PPP:Wan2017Benchmarking} with $454$ images, and similarly select our data as above.
	
	\subsection{Evaluation on Synthetic Data} \label{sec:toy_exp_and_perform_analysis}
	
	We conduct experiments with synthetic data to analyse our method in detail.
	Since the self-supervision (SS) is a major technical component of our method, we study its performance and optimal parameter settings.
	
	First, we analyze the distance loss in Equ.~\eqref{equ:loss_distance_of_latent}. Results are shown in Fig.~\ref{fig:performance_analysis}~(b), where $\text{loss}=1-d_\psi$ and distance indicates $|a - b|$.
	The curves show that the loss values, ranging from $0$ to $1$, indeed decline with increasing distance values but with different shapes.
	According to the curve shapes, we set $\alpha = 1.4$ empirically in the follow experiments.
	
	We then study the balance between the intra-domain distance and inter-domain distance in SS.
	In Equ.~\eqref{equ:ss_object_func}, $\lambda_1$, $\lambda_2$ control the intra-domain distance loss while $\lambda_3$ controls the inter-domain distance loss.
	Therefore, we tune these hyper parameters and re-train the model without touching other settings.
	The resulting mean square errors (MSEs) are shown in the bottom of Fig.~\ref{fig:performance_analysis}~(b), where $(\lambda_1+\lambda_2)/\lambda_3=0.1$ produces the worst model while $(\lambda_1+\lambda_2)/\lambda_3=1$ gives the best one.
	This proves that a good balance should consider both intra-domain and inter-domain distances of the feature, which confirms our Assumption B.
	In practice, we set $\lambda_1=\lambda_2=0.5$ and $\lambda_3=1$ in the following experiment.
	
	Next, we study the impacts of our SS and CC constraints by conducting an ablation study.
	Results are reported in Fig.~\ref{fig:performance_analysis}~(c), showing that removing any of these constraints will increase the final error drastically, \ie, twice as large or even more.
	This persuasively proves that both the SS and CC are essential to our method.
	
	For more information, we also show the results of CycleGAN~\cite{UII:Zhu2017cycleGAN} and UNIT~\cite{UII:Liu2017UnsupervisedImageTranslation} on our synthetic dataset.
	Their MSEs are $0.67/0.61$ and $0.46/0.37$, which are obviously larger than ours ($0.289/0.274$).
	This demonstrates the advantage of our method.
	
	Finally, we consider the overall loss function in Equ.~\eqref{equ:total_object_func}. We introduce two temporary weights  $w_1$ and $w_2$ to balance each term, and the resulting loss function becomes
	\begin{equation} \label{equ:modified_object_func}
	\min_{E, G}\max_{D}  w_1 \mathcal{L}_{SS} + \mathcal{L}_{GAN} + w_2 \mathcal{L}_{CC}.
	\end{equation}
	
\begin{figure} [!htb]
		\centering
		\includegraphics[width=0.8\linewidth]{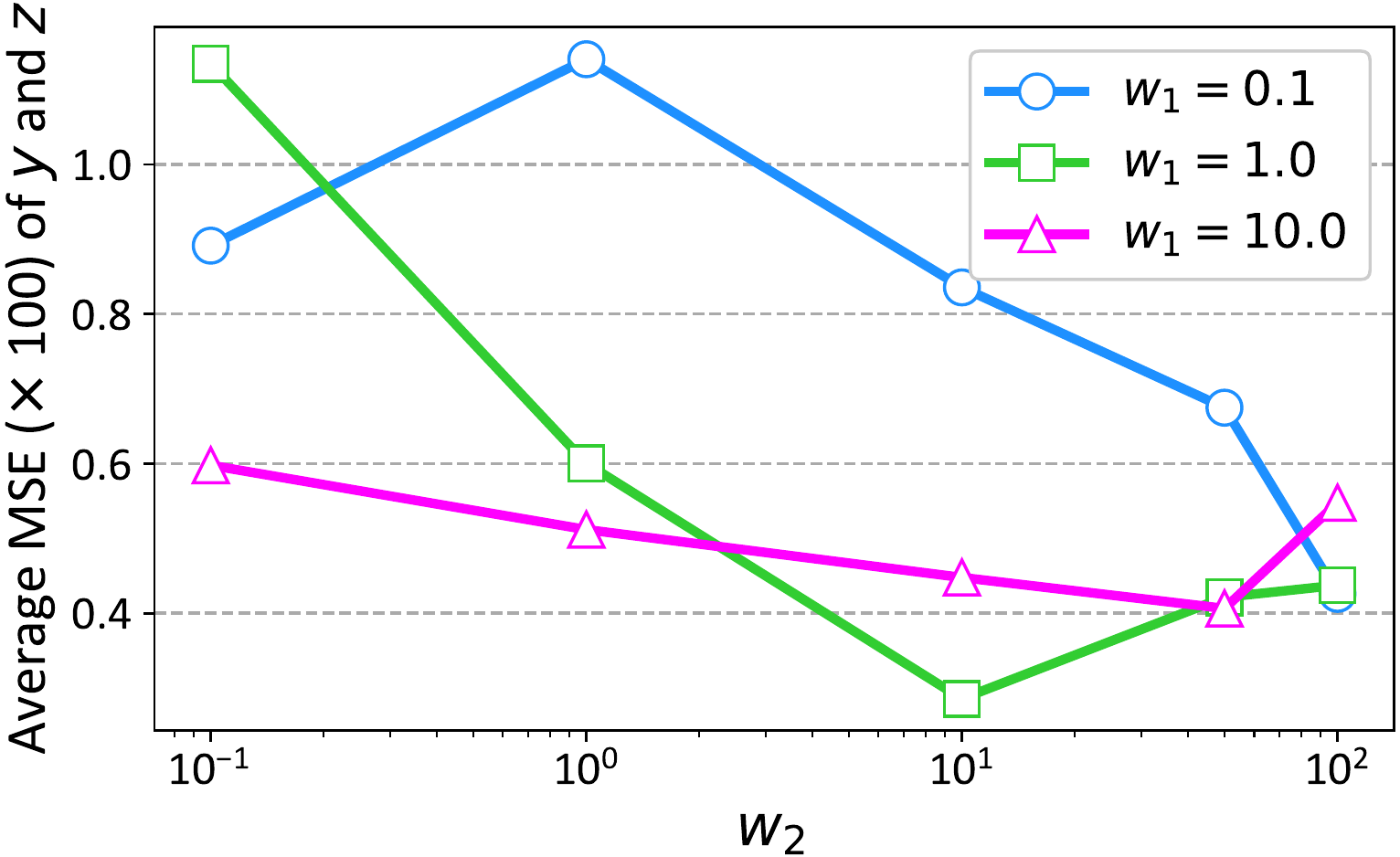}
		\caption{Results with different $w_1$ and $w_2$ in Equ.~\eqref{equ:modified_object_func}.}
		\label{fig:relationship_ss_cc}
	\end{figure}

	We conduct experiments with different $w_1$ and $w_2$ and the results are in Fig.~\ref{fig:relationship_ss_cc}, where we find that $w_1=1$, $w_2=10$ is an optimal choice. According to Equ.~\eqref{equ:consistency_loss_func}, this suggests $\lambda_4 = \lambda_5 =  \lambda_6 = 10.0$ in the experiment.
	
	\begin{figure*} [htbp]
		\centering
		\includegraphics[width=\linewidth]{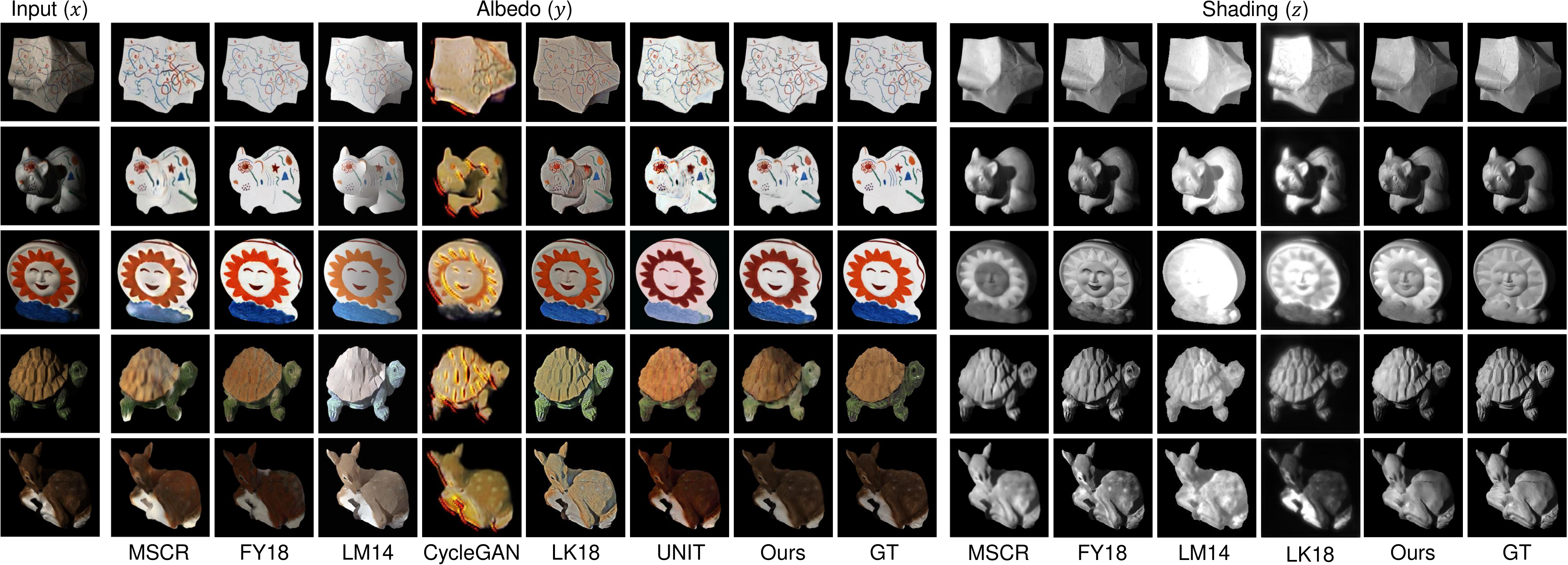}
		\caption{Qualitative comparisons on MIT dataset.
			From left to right: MSCR and FY18 are fully supervised method shown for reference.
			LM14, CycleGAN, LK18 and UINT are unsupervised methods than can be directly compared with outs.
			Our results have a good visual quality and appear almost identical to the ground truth.
		}
		\label{fig:intrinsic_decomposition}
	\end{figure*}

	\subsection{Evaluation on Real Image Separation Tasks}
	\textbf{Intrinsic image decomposition.}
	We compare the proposed method with the most representative unsupervised methods, \ie, LM14~\cite{PPP:Li2014Single}, CycleGAN~\cite{UII:Zhu2017cycleGAN},  UNIT~\cite{UII:Liu2017UnsupervisedImageTranslation} and LK18~\cite{PPP:LVvG18}. They are trained on the same dataset as ours.
	Besides, we also show results of fully-supervised methods MSCR~\cite{PPP:narihira2015direct} and FY18~\cite{PPP:Fan2018revisitingDeepIntrinsic} using their pre-trained models. Note that these fully-supervised methods are trained on ground truth data triplets and thus their results can only serve as a reference.
	
	Visual comparison among all the methods is illustrated in Fig.~\ref{fig:intrinsic_decomposition}.
	CycleGAN produces the worst images with bad textures and wrong colors. Its training may fail with a high probability.
	LK18 and LM14 perform better than CycleGAN while their visual qualities are still unsatisfactory.
	UNIT produces relatively good albedo layers. However, it cannot output the shading layer due to its image-to-image translation framework.
	On the other hand, our method generates both the albedo and shading layers with a good quality and they appear highly identical to the ground truth images. The visual quality of our results is even better than the two fully supervised methods, \ie, MSCR and FY18, in some cases.

	We also provide a qualitative comparison in Table~\ref{tab:numerical_comparisons_mit}, where LMSE is an error metric specifically designed for intrinsic image data introduced by~\cite{PPP:grosse2009groundLMSE}.
	Numerical results show that our method achieves state-of-the-art performance among these unsupervised methods.
	Its accuracy is even comparable to the fully supervised methods that rely on exact ground truth data triplets in training.

	\begin{table}[htbp]
		\begin{center}
			\setlength{\tabcolsep}{1.5mm}
			\begin{tabular}{cccccc}
				\toprule[1.3pt]
				
				& \multicolumn{ 1}{l}{} & \multicolumn{ 3}{c}{MSE} & LMSE \\
				& \multicolumn{ 1}{l}{Method} &                             Albedo &     Shading &          Avg. &       Total  \\
				\hline
				\specialrule{0em}{1pt}{1pt}
				
				\multirow{4}{*}{\rotatebox{90}{Unsup.}}
				
				& LM14 	&					       0.0286 &      0.0227  &       0.0255 &     0.0366 \\
				& CycleGAN & 		               0.0272 &        N/A     &       0.0272 &    0.407      \\
				& UNIT  &   0.0207 &        N/A     &    0.0207    &   0.0310     \\
				& LK18  &   0.0232 &     0.0166     &    0.0197    &   0.0379     \\
				&         Ours &                                     {\bf 0.0167} &  {\bf 0.0140}    &   {\bf 0.0154}&{\bf 0.025} \\
				
				\hline
				\specialrule{0em}{1pt}{1pt}
				\multirow{2}{*}{\rotatebox{90}{\tabincell{c}{Sup.\\ (Ref.)}}}
				
				& MSCR &                     0.0207 &   0.0124   &        0.0165 &    0.024  \\
				& FY18 &         0.0127 &   0.0085  &        0.0106 &    0.020  \\

				\bottomrule[1.3pt]
			\end{tabular}
		\end{center}
		\caption{Numerical comparison on MIT intrinsic dataset. Note that MSCR and FY18 are fully-supervised methods, which are trained on ground truth data triplets, and thus their results can only serve as a reference.}
		\label{tab:numerical_comparisons_mit}
	\end{table}
	
\begin{figure} [htb]
		\centering
		\includegraphics[width=0.9\linewidth]{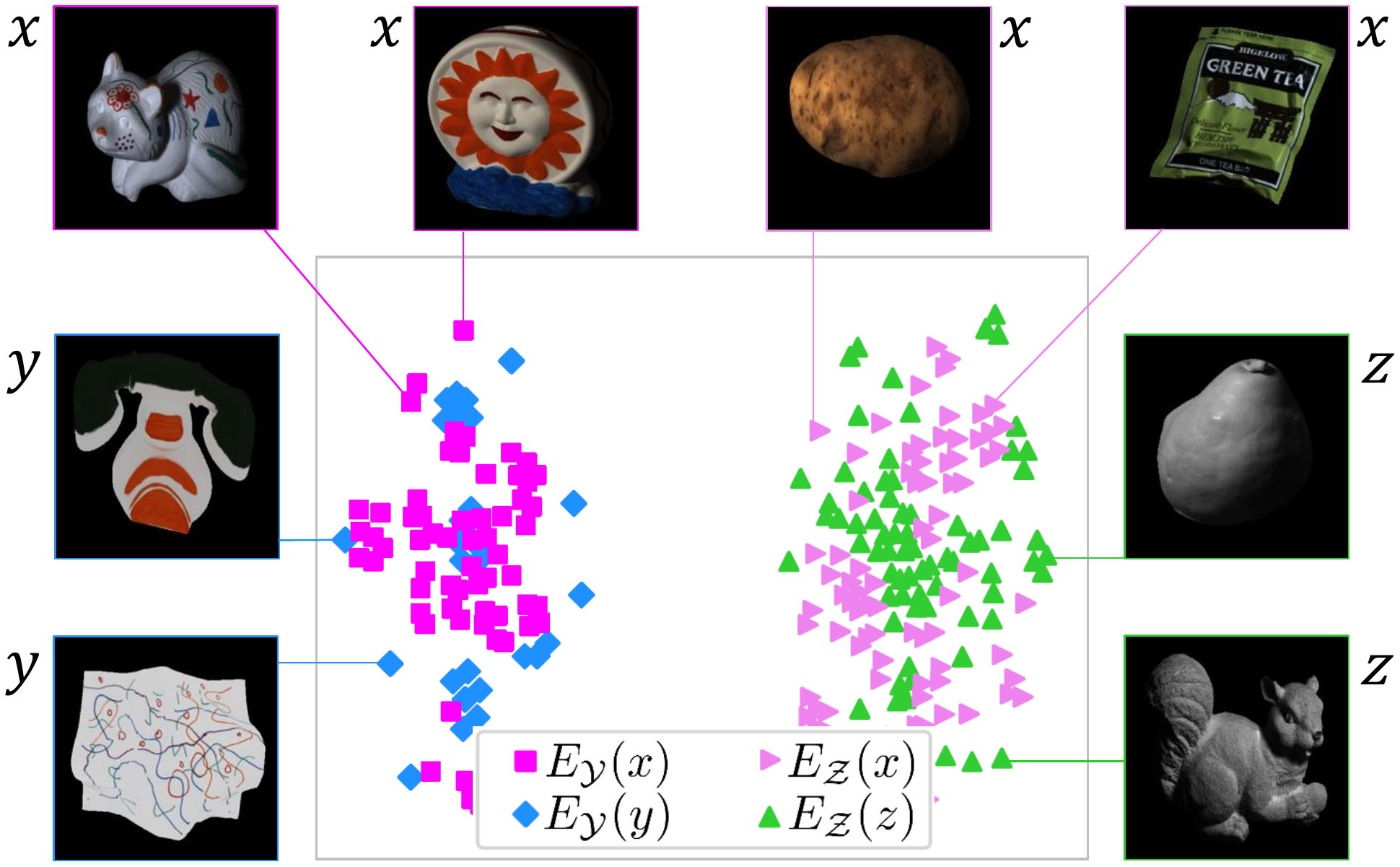}
		\caption{Visualization of our feature encoding distribution using the corresponding PCA vectors.
			We use t-SNE~\cite{PPP:Hinton2008tSNE} to aid the visualization of the latent space.
			Encoder $E_{\mathcal{Y}}$ projects $x$ and $y$ into the same cluster (latent space for albedo layer), and encoder $E_{\mathcal{Z}}$ projects $x$ and $z$ into another cluster (latent space for shading layer).
		}
		\label{fig:cluster_intrinsic}
	\end{figure}

	In this experiment, we also study our feature extraction results.
	Fig.~\ref{fig:cluster_intrinsic} visualizes the distribution of the encoded features by using their $2$D PCA vectors.
	It is clear that the encoder $E_{\mathcal{Y}}$ projects $x$ and $y$ into the same cluster (indicating the same latent space for albedo layer), and the encoder $E_{\mathcal{Z}}$ projects $x$ and $z$ into another cluster (latent space for shading layer). The two clusters stay apart from each other with a large distance. This verifies the correctness of our Assumption A and B and thus confirms the design of our method.

	\begin{figure*} [!htb]
		\centering
		\includegraphics[width=0.72\linewidth]{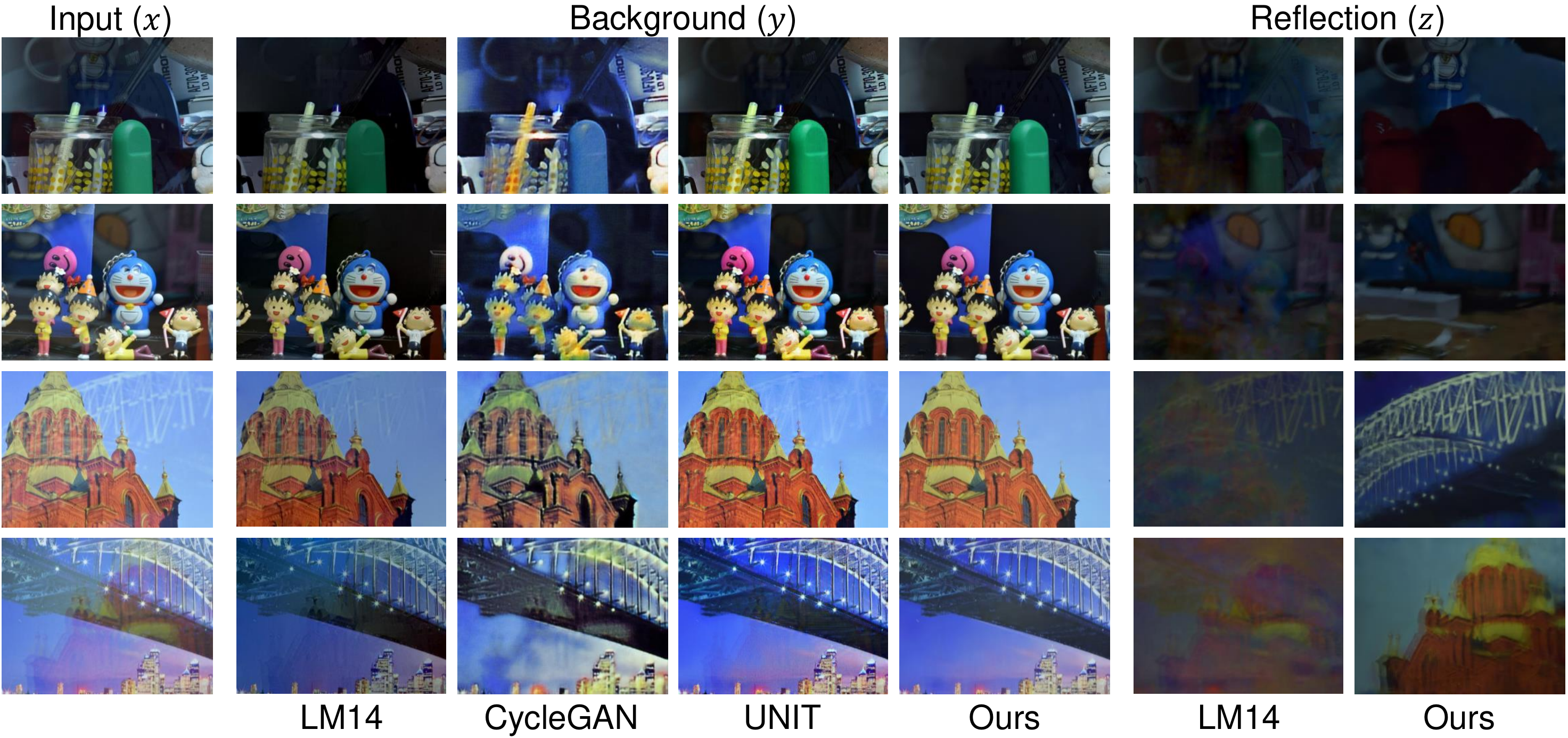}
		\caption{Visual results by different unsupervised methods for reflection removal.
			From left to right, LM14,
			CycleGAN and UNIT.
			Note that CycleGAN and UINT only recover background image without producing the reflection image.
		}
		\label{fig:reflection_removal}
	\end{figure*}

	\textbf{Reflection removal.}
	Similar to the intrinsic image decomposition case, we compare our method with the unsupervised methods LM14, CycleGAN and UNIT.
	They are trained on the same dataset.
	For the fully-supervised methods, we use the pre-trained models of ZN18~\cite{PPP:Zhang2018PerceptualLosses} and BDN~\cite{PPP:Yang18BDN} on the same dataset but with ground truth triplet data.
	Fig.~\ref{fig:reflection_removal} gives the visual comparison among unsupervised methods and ours. LM14 can separate both background and reflection layers from the input image, but the separation is clearly not ideal. Both the CycleGAN and UNIT only recover the background layer, where there still exist artifacts or remaining reflections in many image regions. Overall, our method achieves the best background cleanness and the visual quality of the reflection layer.

	To make a qualitative comparison with state-of-the-art methods, we use peak-signal-to-noise-ratio (PSNR)~\cite{PPP:huynh2008scopePSNR} and structural similarity index (SSIM)~\cite{PPP:wang2004imageSSIM} as
	evaluation metrics.
	Table~\ref{tab:numerical_comparisons_ben} shows that our method achieves the state-of-the-art performance against other unsupervised methods, and is also superior to the fully-supervised method BDN~\cite{PPP:Yang18BDN}.

	\begin{table}
		\begin{center}
			\setlength{\tabcolsep}{2.1mm}
			\begin{tabular}{cccccc}
				\toprule[1.3pt]
				
				& \multicolumn{ 1}{l}{} & \multicolumn{ 2}{c}{Background} & \multicolumn{ 2}{c}{Reflection} \\
				& \multicolumn{ 1}{l}{Method} &                              SSIM &       PSNR &          SSIM &       PSNR  \\
				\hline
				\specialrule{0em}{1pt}{1pt}
				
				\multirow{4}{*}{\rotatebox{90}{Unsup.}}
				
				& LM14~ 	&					       0.715 &      17.00 &      0.357 &      17.58 \\
				& CycleGAN & 		               0.622 &      16.63     &       N/A &    N/A      \\
				& UNIT  &    0.738 &      19.23  &  N/A          &   N/A         \\
				&         Ours &                                     {\bf 0.821} &  {\bf 20.47}    &   {\bf 0.425}&{\bf 19.31} \\
				
				\hline
				\specialrule{0em}{1pt}{1pt}
				\multirow{2}{*}{\rotatebox{90}{\tabincell{c}{Sup.\\ (Ref.)}}}
				
				& BDN &                                                   0.820 &    18.87   &           N/A &      N/A  \\
				& ZN18 &                0.858 &      22.22 &         0.420 &       18.94 \\

				\bottomrule[1.3pt]
			\end{tabular}
		\end{center}
		\caption{Comparison on reflection removal dataset.
			Note that BDN and ZN18 are fully-supervised methods, which are trained on ground truth data triplets, and thus their results can only serve as a reference. }
		\label{tab:numerical_comparisons_ben}
	\end{table}
	
	\subsection{Evaluation on Multi-layer Separation}
	
	\begin{figure} [!htb]
		\centering
		\begin{tabular}{cc}
			\includegraphics[width=0.98\linewidth]{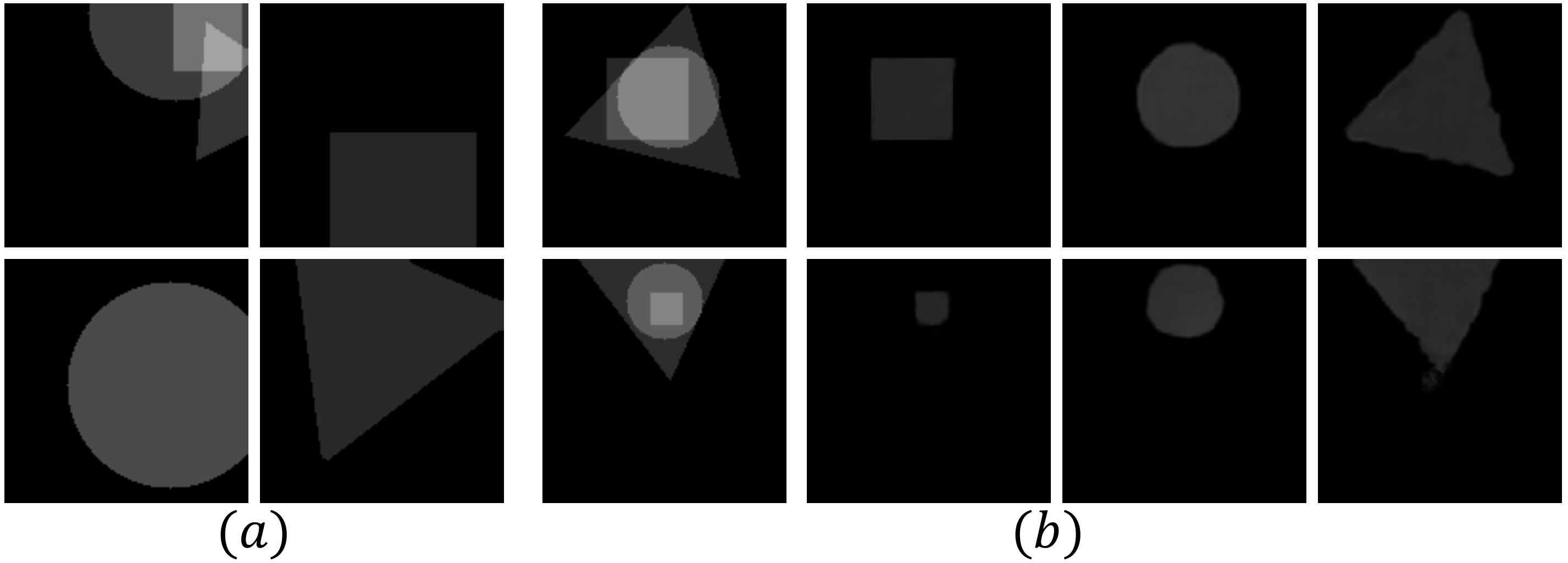}
		\end{tabular}
		\caption{The extend experiment on three layers separation task.
			(a) Example of the blend image and each layer.
			(b) Two visual results obtained by the proposed method.
			From left to right: input image and separated layers.}
		\label{fig:split_three_layers}
	\end{figure}
	
	In this section, we extend our evaluation to the multi-layer case. We re-render the synthetic dataset with three different shapes, \ie, square, circle and triangle, as shown in Fig.~\ref{fig:split_three_layers} (a). Totally $5000$ images are synthesized without containing any triplet data, and other settings remain the same as described in Sec.~\ref{sec:toy_exp_and_perform_analysis}. Accordingly, we modify our method by adding another network branch of a generator and a discriminator to handle the third layer.
	The example results are shown in Fig.~\ref{fig:split_three_layers} (b), which demonstrate the successfulness of our method to handle such a challenging problem.

	\section{Conclusion}
	
	In this paper, we propose an unsupervised single image layer separation method.
	It is unsupervised since it requires no ground truth data triplet, \ie, input images and its layer images for training.
	To this end, we introduce two assumptions about the feature distributions of different layers, namely the shared information consistency and layer independence assumptions. Based on them, we design our method by using the GANs framework with self-supervision and cycle consistency constraints.
	Experimental results show that our method outperforms existing unsupervised methods in both synthetic and real world tasks. The method can also be extended to solve a more challenging three-layer separation task.
	
	\newpage
	\bibliography{egbib}
	\bibliographystyle{aaai}
	
\end{document}